\documentclass[letterpaper, 10 pt, conference]{ieeeconf}

\IEEEoverridecommandlockouts 
\usepackage{tikz}
\usetikzlibrary{shapes,shadows,arrows}
\usepackage{hyperref}
\usepackage{listings}
\usepackage{color}
 
\definecolor{codegreen}{rgb}{0,0.6,0}
\definecolor{codegray}{rgb}{0.5,0.5,0.5}
\definecolor{codepurple}{rgb}{0.58,0,0.82}
\definecolor{backcolour}{rgb}{0.95,0.95,0.92}
 
\lstdefinestyle{mystyle}{
    backgroundcolor=\color{backcolour},   
    commentstyle=\color{codegreen},
    keywordstyle=\color{magenta},
    numberstyle=\tiny\color{codegray},
    stringstyle=\color{codepurple},
    basicstyle=\scriptsize,
    breakatwhitespace=false,         
    breaklines=true,                 
    captionpos=b,                    
    keepspaces=true,                 
    numbers=left,                    
    numbersep=5pt,                  
    showspaces=false,                
    showstringspaces=false,
    showtabs=false,                  
    tabsize=2
}
\newcommand{\code}[1]{\texttt{\small{#1}}}

\newcommand\barrier[2][]{
\def\x{10}
  \begin{scope}[shift={(#2)}]
    \draw[ultra thick,gray] (0/\x,1/\x) -- (1/\x,0/\x);
    \draw[ultra thick,gray] (1/\x,1/\x) -- (2/\x,0/\x);
    \draw[ultra thick,gray] (2/\x,1/\x) -- (3/\x,0/\x);
    \draw[ultra thick,gray] (3/\x,1/\x) -- (4/\x,0/\x);
    \draw[ultra thick,gray] (4/\x,1/\x) -- (5/\x,0/\x);
    \draw[thick,gray] (1.25/\x,-0.5/\x) -- (1.25/\x,-2.5/\x);
    \draw[ultra thick,gray] (0.75/\x,-2.5/\x) -- (1.75/\x,-2.5/\x);
    \draw[thick,gray] (3.75/\x,-0.5/\x) -- (3.75/\x,-2.5/\x);
    \draw[ultra thick,gray] (3.25/\x,-2.5/\x) -- (4.25/\x,-2.5/\x);
  \end{scope}
}

\tikzstyle{decision} = [diamond, fill=red!50, draw]
\tikzstyle{line} = [draw, -stealth, thick]
\tikzstyle{elli}=[draw, ellipse, fill=red!50,minimum height=4mm,
text width=6em, text centered]
\tikzstyle{block} = [draw, rectangle, fill=blue!50, text
width=4.5em, text centered, minimum height=2.5mm, node
distance=6em]

\title{\LARGE \bf
ROS and Buzz: consensus-based behaviors for heterogeneous teams
}

\author{David~St-Onge,Vivek~Shankar~Varadharajan, Guannan Li,
Ivan Svogor and Giovanni~Beltrame
\thanks{Dr. St-Onge, M. Varadharajan, M. Li, Dr. Svogor and Dr.
Beltrame are with the Department
of Computer and Software Engineering, \'Ecole Polytechnique de
Montr\'eal, 2900 Boul \'Edouard-Montpetit, Qu\'ebec
CA e-mail: (david.st-onge@polymtl.ca).}}
\begin{document}

\maketitle
\thispagestyle{empty}
\pagestyle{empty}

\begin{abstract}

  This paper address the challenges encountered by developers when deploying a
  distributed decision-making behavior on heterogeneous robotic systems. Many
  applications benefit from the use of multiple robots, but their scalability
  and applicability are fundamentally limited if relying on a central control
  station.  Getting beyond the centralized approach can increase the
  complexity of the embedded intelligence, the sensitivity to the network
  topology, and render the deployment on physical robots tedious and error-prone. By integrating the swarm-oriented programming language Buzz with the
  standard environment of ROS, this work demonstrates that behaviors requiring
  distributed consensus can be successfully deployed in practice. From
  simulation to the field, the behavioral script stays untouched and
  applicable to heterogeneous robot teams. We present the software
  structure of our solution as well as the swarm-oriented paradigms required
  from Buzz to implement a robust generic consensus strategy. We show the
  applicability of our solution with simulations and experiments with
  heterogeneous ground-and-air robotic teams.
\end{abstract}

\section{INTRODUCTION}

The range of applications for multi-robot systems is constantly
and rapidly expanding. Small groups of heterogeneous robots
collaborating to extend their individual potential was
repeatedly proven to be successful~\cite{Dudek2000}.
Nevertheless, each complex unit of these scenarios are mandatory
and a single failure will most likely make the mission fail. By
leveraging a greater number of similar agents, individual
failure can be compensated and sensor imprecision can be
mitigated by merging many sources. Swarm intelligence is known
for decades to be a solution to many complex problems in
dynamic, hostile and unknown environment. Any robotic solution
designed for this purpose requires to be flexible, scalable and
robust, the genuine definition of Swarm Robotics Systems
(SRS)~\cite{Sahin2004}. Unfortunately, tools to ease their
implementation are hardly available.

Researchers are very active on developing behaviors for
robotic swarms supported by companies providing the required
hardware, such as the Kheperas~\cite{kteam} and some open source
initiatives such as the Zooids~\cite{Goc2016}. These affordable
platforms grant access to physical implementation with
significant number of robots, but lack a common set of software
tools for their programming. Indeed, all swarms have a lot in common. Swarm members are all decentralized systems without predefined roles, based on simple local interaction implemented through techniques like situated communication, neighbor management, shared environment--based data, etc. For a swarm system, in particularly heterogeneous swarm, these currently need to be re-implemented for each platform and experiment.


\begin{figure}[!h]
\begin{center}
  \includegraphics[width=0.4\textwidth]{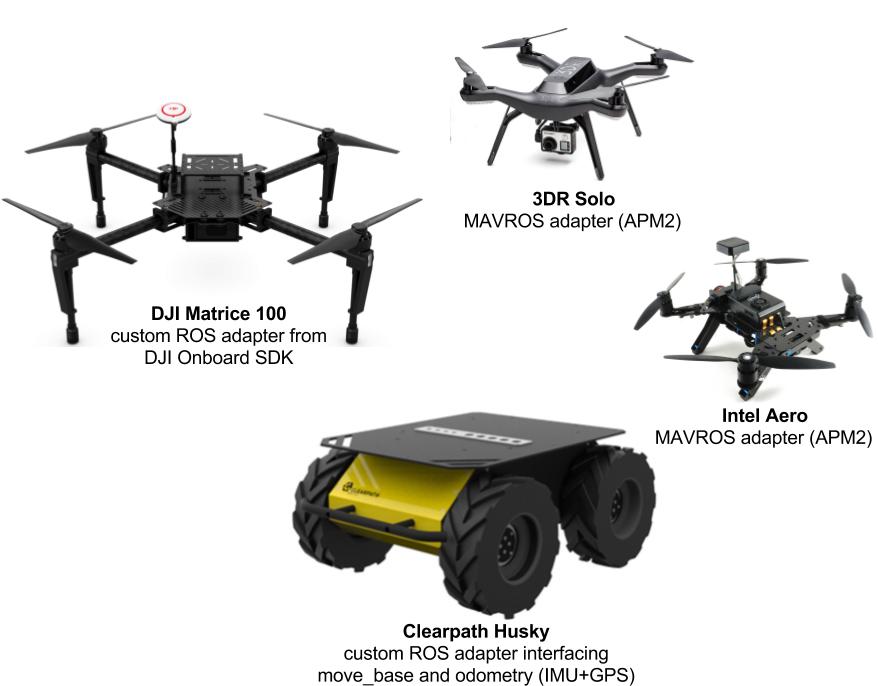}
\caption{Platforms tested with ROSBuzz.}
\label{fig:heterogenous}       
\end{center}
\end{figure}

Within the heterogeneous swarm context, creation of an optimized and specialized software infrastructure flexible enough to make robotics researchers feel unconstrained, while simultaneously increasing their development efficiency is a tedious task. One that is addressed with ROS for single robot, but in the context of a swarm this needs to be explored further. The need became more apparent with the appearance of programming languages specific for swarm development, e.g. Voltron and Proto. However, due to their high level of abstraction for addressing swarm members, in 2016 MIST Laboratory released Buzz; a (domain-specific) programming language specific to robot swarms~\cite{Pinciroli2015}. Its purpose is to help researchers and practitioners with swarm software development by providing a set of primitives which accelerates the implementation of swarm-specific behaviors. Buzz comes with an optimized virtual machine for distributed systems and deals with each specific swarm member, and each specific swarm member executes the same script. This  bottom--up behavior deployment based on abilities of each robot in the swarm is merged with the top--down swarm strategy of the whole group. Such heterogeneity allows a script to be deployed on any autonomous robots from the small desk robots, to UAVs and UGVs of any size, and even satellites. While Buzz is natively deployed on Kilobots~\cite{kteam} and Kheperas, larger robots require integration within a software ecosystem that will allow roboticists to interface with different sensors, actuators and complex algorithms.

To address issue we introduce ROSBuzz, the ROS implementation of the BVM. Much more than another software wrapper, it enables for a) fast script--based programming of complex behaviors, b) seamless script porting on different hardware, and most importantly c) it allows for coherent performance from simulation to field deployment.
\noindent To present ROSBuzz we first describe the key primitives of Buzz, explain the details of its software architecture, provide detailed explanation of behavior robustness through distributed intelligence in Buzz and exemplify the distributed consensus strategy used in ROSBuzz. Finally, we demonstrate ROSBuzz by deploying it on a heterogeneous swarm (consisting of some of the hardware shown in Fig.~\ref{fig:heterogenous}) in the simulation environment (software in the loop) and real world experiment.


\section{Related Work}
\label{related work}

Swarms of UAVs are challenging to implement, but their high potential
to be robust, resilient and flexible~\cite{Brambilla2013} motivates a
number of robotics laboratories. For instance, the Ecole Polytechnique Federale 
de Lausanne Laboratory of Intelligent
Systems~\cite{Hauert2011} introduced fixed-wing UAVs to
demonstrate flocking~\cite{reynolds1987flocks} with platform specific 
programming.
Flocking is part of basic swarm behaviors that do not require formal 
consensus over the group~\cite{Bayindir2016}.

The Johns Hopkins University Applied Physics
Laboratory~\cite{Bamberger2006} design fixed wing planes to demonstrate a set 
of swarm concepts for UAVs
(e.g. aggregation) using a shared consensus variable without the need for global 
communication,
 but they did not develop a robot-agnostic platform.
 
Consensus-based approaches have been proposed for a
number of multi-UAV coordination problems such as resource and task allocation 
\cite{Brunet2008}, formation control~\cite{WeiRen2005},~\cite{Kuriki2015}, and 
determination of coordination variables~\cite{Wei2015}. Each approaches are specific to their application and hardware 
implementation.

The Naval Postgraduate School Advanced Robotic Systems Engineering
Laboratory~\cite{Davis2016,chung201350} extend this idea and develop a
software infrastructure for a fleet of UAVs with ROS-based on-board
computers connected to a ground station through the MAVlink
protocol~\cite{mavlinkweb}.
The UAVs are loaded with multiple behavior binaries before launch, and
the user can then activate and deactivate them while in flight. The
authors conducted flights with 50 fixed-wing UAVs to test the
infrastructure~\cite{Chung2016}. These
works~\cite{Davis2016,Chung2016} do not integrate a consensus mechanism to agree
on the behavior to be used. However, they do present different
consensus algorithms for the coordination within the
swarm.

In an effort to standardize the swarm programming, Georgia Tech created the 
Robotarium, 
to test swarm behaviors remotely with desk robots~\cite{Pickem2016}. Their API 
is restricted to
the specific custom robots of their system and do not include a generic 
consensus strategy.

\section{Buzz in ROS}
\subsection{Buzz}

In order to accelerate the implementation of swarm behaviors, Buzz provides three important primitives: a) virtual stigmergy, b) swarm aggregation and c) neighbor interaction. 

\textit{Virtual Stigmergy} is a software implementation of bio--inspired stigmergy, which uses the environment for coordination between swarm members i.e. an environment-mediated communication modality~\cite{camazine2002self}.
\noindent Virtual stigmergy is implemented as a shared memory table containing $\langle$\code{key},\code{value}$\rangle$ pairs distinguishable through a unique identifier. Shared memory table is managed among the robots by maintaining a local copy of the table on each robot which is synchronized by exchanging metadata. The metadata contains a) a Lamport clock~\cite{lamport1978time} which increments on each modification and b) the ID of the last robot that modified the data. This information is broadcasted between swarm members, and the decision whether or not to re--broadcast is inferred by analyzing the Lamport clock. For the possible conflicts due to concurrent modification, the BVM has a set of rules for conflict resolution. More on this can be found in~\cite{Pinciroli2016}.

\textit{Swarm Aggregation} is a primitive which allows for grouping of robots into sub--swarms, through the principle of dynamic labeling~\cite{Pinciroli2016b}. The \code{swarm}
construct is used to create a group of robots which can be attributed with a specific behavior, which slightly differs from the behavior of the remaining swarm, based either on the task or robot abilities. 
\textit{Neighbors Operations} in Buzz refer to a rich set
of functions which can be performed with or on neighboring robots through the situated communication~\cite{Støy2001} principle. Neighbors are defined from a network--based perspective as robots which have a direct communication link with each other. With situated communication, whenever a robot receives a message, the origin position of the message is known to the receiver. With this, one can obtain range and bearing information and avoid the use of global positioning. However, for more precise movement, as in our experiment, instead of range and bearing, messages between robots use GPS coordinates. 
Finally, the Buzz script is compiled into an optimized, space efficient, and platform-agnostic bytecode to be executed on the BVM. In order to interface the BVM with the remaining specific actuators and sensors, we use ROS. The following section describes how are Buzz and ROS integrated together, to allow seamless and platform-agnostic Buzz extension and execution.

\subsection{ROSBuzz}
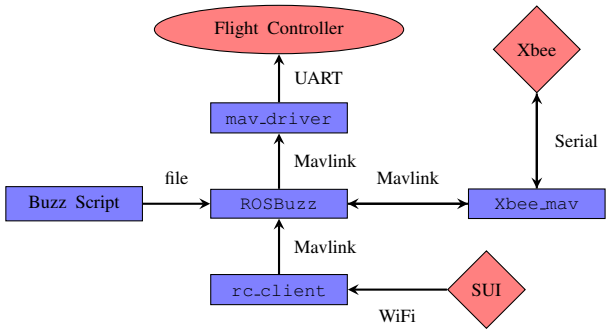
\begin{figure}[!h]\centering
\begin{tikzpicture}[scale=0.8]
\node [block] (driver) {\scriptsize {\tt mav\_driver}};
\node [elli, above of=driver, yshift=0.5em] (FC) {\scriptsize Flight
Controller};
\node [block, below of=driver, yshift=2.75em] (ROSBuzz) {\scriptsize{\tt
ROSBuzz } };
\node [block, left of=ROSBuzz, xshift=-1.75em] (script) {\scriptsize
Buzz Script};
\node [block, right of=ROSBuzz, xshift=3.75em] (xbeemav) {\scriptsize
{\tt Xbee\_mav}};
\node [block, below of=ROSBuzz, yshift=2.75em] (rc) {\scriptsize{\tt
rc\_client}};
\node[decision, above of=xbeemav, yshift=3em](Xbee){\scriptsize
Xbee};
\node[decision, right of=rc, xshift=5em](UI){\scriptsize SUI};
\path [line] (driver) -- node[yshift=0em, xshift=1.5em] {\scriptsize
UART} (FC);
\path [line] (ROSBuzz) -- node[yshift=0em, xshift=1.75em]
{\scriptsize Mavlink} (driver);
\path [line] (script) -- node[yshift=1em, xshift=0em] {\scriptsize
file}(ROSBuzz);
\path [line] (rc) -- node[yshift=0em, xshift=1.75em] {\scriptsize
Mavlink}(ROSBuzz);
\path [line] (ROSBuzz) -- node[yshift=1em, xshift=0em] {\scriptsize
Mavlink}(xbeemav);
\path [line] (xbeemav) -- (ROSBuzz);
\path [line] (UI) -- node[yshift=-1em, xshift=0em] {\scriptsize WiFi}
(rc);
\path [line] (xbeemav) -- node[yshift=0em, xshift=1.5em] {\scriptsize
Serial} (Xbee);
\path [line] (Xbee) -- (xbeemav);
\end{tikzpicture}
\caption{Overview of the different on-board modules required
around ROSBuzz.}
\label{fig:struc}
\end{figure}

The ROS implementation of Buzz was originally driven by the need
to port swarm behaviors to an heterogeneous fleet of UAVs. In
order to maximize the compatibility of the node to various
platforms, the input and output messages and services follow
MAVlink protocol using its MAVROS implementation. To compile the
node, the open source Buzz library must be available on the
system as well as \code{geometry\_msgs}, \code{std\_msgs} and 
\code{mavros\_msgs}. Fig. \ref{fig:struc} shows the ecosystem of a
minimal ROSBuzz deployment. The serialized and optimized Buzz
messages payloads are transferred through a MAVlink standard
payload message, transmitted for instance by a Xbee
communication module (\code{xbeemav}). The Buzz virtual machine takes a script 
as input, specified by the user in its launch file, and loop on its
{\tt step} function.

\begin{figure}[h!]	
    \centering
    \includegraphics[width=\linewidth]{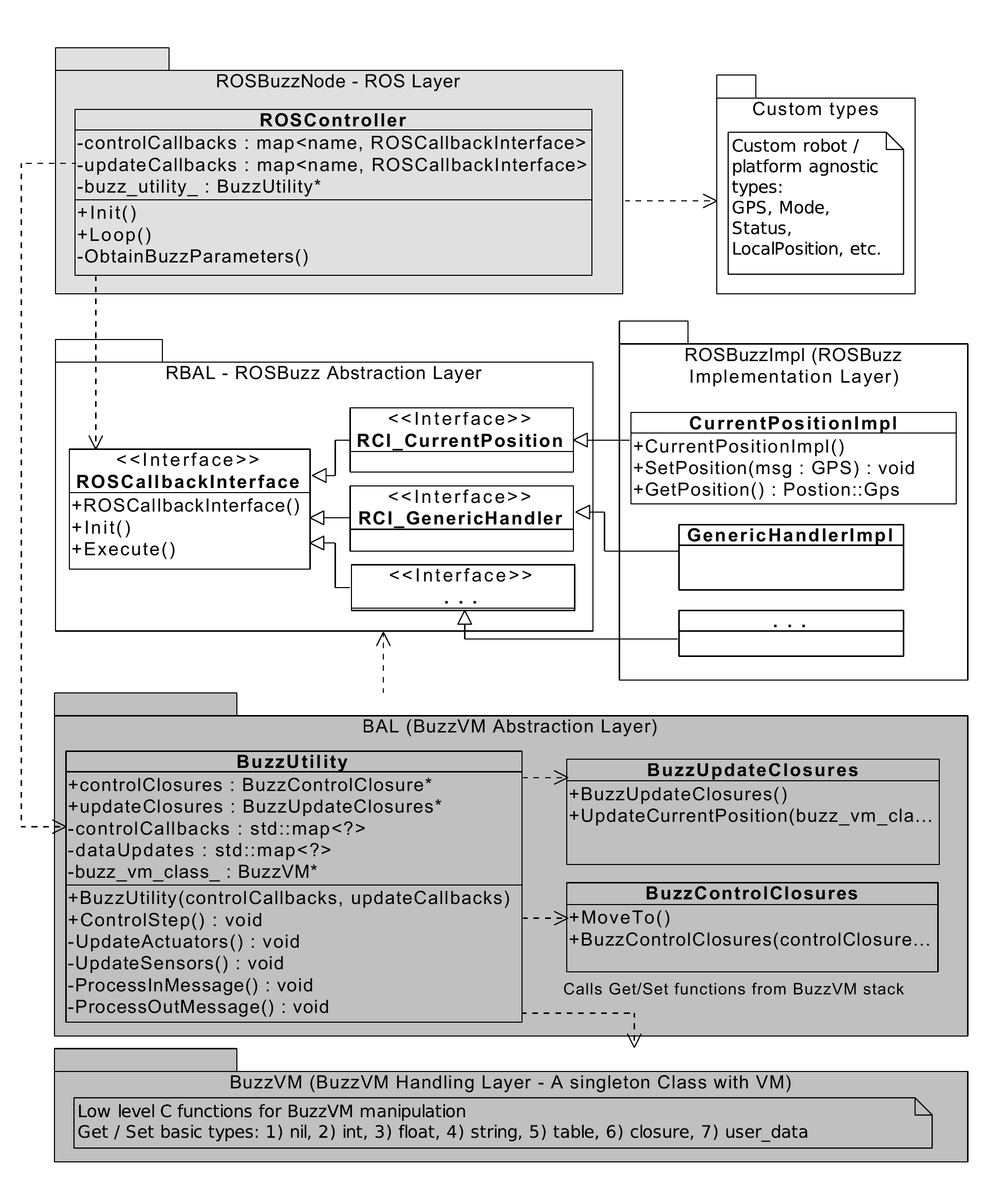}
    \caption{Simplified UML Class diagram of the ROSBuzz software architecture}
	\label{fig:rosbuzz-uml}
\end{figure}

The software architecture of ROSBuzz shown in Fig. \ref{fig:rosbuzz-uml} is 
organized in four distinct layers which reconcile the step--based execution 
nature of Buzz and the event--based nature of ROS.

The top most \textit{ROS Layer}, is essentially a ROS node with a role to 
initialize all the necessary environmental parameters and to start the main ROS 
loop. The most important initialization parameters are the collections of 
callback functions, i.e. \code{updateCallbacks} and \code{controlCalbacks} 
which hold references to implementations of robot specific operations for 
sensing (the former) and actuation (the latter). Since we are dealing with a 
heterogeneous swarm, there can always be slight differences in the ROS topic 
naming, data types, procedures (e.g. with drones, the takeoff or landing 
procedures). For this purpose, the callback functions are implemented on a robot 
type basis through the \textit{ROS Abstraction Layer}. With this, each specific 
robot operation is introduced to ROSBuzz as a module with a standardized 
\code{ROSCallbackInterface}. This ensures that the lower, Buzz Abstraction 
Layer is completely independent of the implementation details since all modules 
must be executable through the inherited \code{Execute} function. 

\noindent In that way, BuzzVM Abstraction Layer acts as a mediator between ROS 
(event driven) and Buzz (step driven) so that in each step of the main ROS loop 
(\code{ROSController} object), an instance of a \code{BuzzUtility} class is 
used to perform the following operations: a) process incoming 
messages, b) update sensors information, c) perform a control step, d) 
process outgoing messages and finally e) update the actuation commands. Operations a) and 
d) refer to information exchange between swarm members, operations b) and e) 
refer to data exchange between ROS and Buzz while operation and c) refers to 
the execution of a Buzz script. For Buzz to propagate the actuation or update 
the state of the robot it uses closure functions\footnote{In this context 
it refers to C functions registered in BVM, available for usage within Buzz scripts}, 
or more specifically it uses instances of the \code{BuzzUpdateClosures} and 
\code{BuzzControlClosures}. The former is used to push and obtain information 
from BuzzVM, while the latter is used to perform actuation from Buzz by 
triggering ROS control callback functions. 

With the described software architecture abstractions, ROSBuzz is easy to manage, 
maintain, upgrade, and most importantly, it simplifies the introduction of new 
robots (of different types) into an existing swarm, with all swarm members 
executing the exact same Buzz control script. 


\subsection{Simulator}
Due to usage of sensitive and expensive flying hardware, simulation is mandatory. The BVM was integrated from the initial development phase in the ARGoS 
simulator~\cite{Pinciroli2012} seamlessly and a QT editor allows to iterate quickly 
in the development of the behavioral script. Unfortunately, it lacks accuracy 
in the simulated dynamics of the robots and is not compliant with a ROS 
architecture. Thus, another simulating environment was developed based on 
Gazebo and leveraging community packages available for ROS. Therefor, the three adapters 
of Fig.~\ref{fig:heterogenous} (DJI, Husky and MAVROS) were also derived for 
Gazebo using the hector package~\cite{Meyer2012} for the Matrices, the 
DroneKit-SITL for the Solos and the nodes provided by Clearpath for the Husky. 
As for the inter-robot communication, to simulate the node \code{xbeemav} (Fig.~\ref{fig:struc}), we implemented a relay node. It transmits the messages out of one instance of ROSBuzz in the simulation to the other instances running in parallel. A 
Bernouilli distribution simulates a given packet drop probability.

\section{Consensus strategy}

When dealing with the coordination of multiple robots, the convergence of all robots to a 
common value regarding their state or their knowledge of the environment, is 
called \emph{consensus}. In Buzz, an include file manages all the required logic 
for the implementation of a barrier to consensus. The barrier is a direct 
implementation of consensus among the robots in a swarm. For its implementation we used virtual stigmergy and swarm table. Each robot updates a value associated to its ID on the shared table and consensus is reached when this table size equals the swarm size. This barrier state also halts further behavior until a global consensus is reached. The Buzz functions are detailed in the 
snippet \ref{lst:barrier}.

\begin{lstlisting}[language=Python,label=lst:barrier,caption=Barrier 
implementation in Buzz.]
BARRIER_VSTIG = 0
BARRIER_TIMEOUT = 600
# Create the barrier
function barrier_create() {
  # reset the timeout counter
  timeIn = 0
  # create the barrier virtual stigmergy
  if (barrier != nil) {
    barrier = nil
    BARRIER_VSTIG = BARRIER_VSTIG +1
  }
  barrier = stigmergy.create(BARRIER_VSTIG)
}

# Executes the barrier
function barrier_wait(threshold, transf, resumef) {
  # share that you are in the barrier
  barrier.put(id, 1)
  # look for the stigmergy status
  barrier.get(id)
  if(barrier.size() - 1 >= threshold or barrier.get("d") == 1) {
    # Going out. Share the barrier is done
    barrier.put("d", 1)
    timeW = 0
    # launch next state
    transf()
  } else if(timeW >= BARRIER_TIMEOUT) {
    # timed out
    barrier = nil
    timeW = 0
    # launch safe resume state
    resumef()
  }
  timeW = timeW+1
}
\end{lstlisting}

\subsection{Simulations}
\begin{figure}[!h]
\begin{center}
  \includegraphics[width=0.45\textwidth]{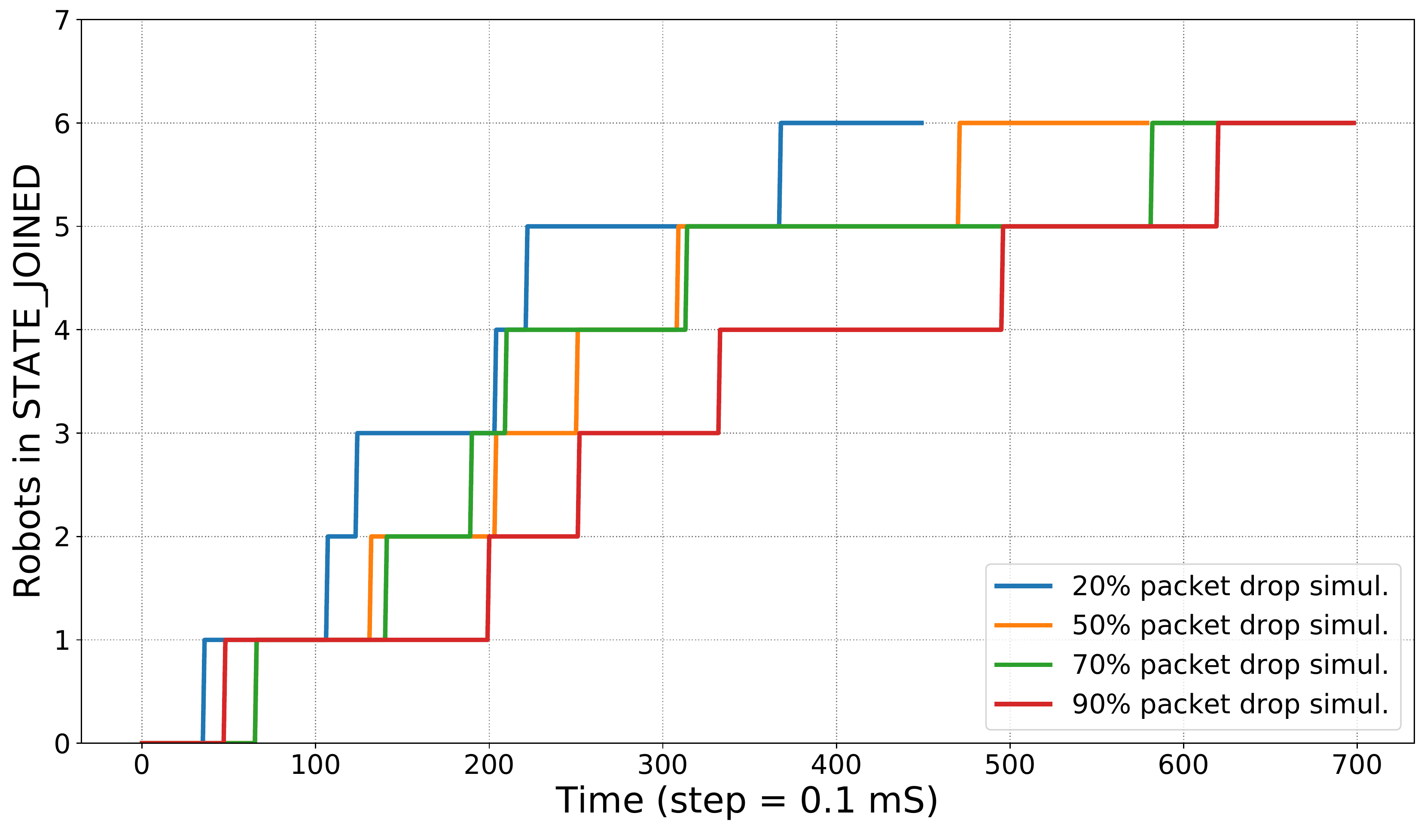}
\caption{Time required for six robots to join in simulation following different packet drop rates.}
\label{fig:joinsim}       
\end{center}
\end{figure}

To test the converge of the barrier and its robustness to various 
communication interference, a set of simulations were conducted with six DJI Matrice 
100.
The results in Fig.~\ref{fig:joinsim} illustrate the time required by each 
robot to reach their task following the exact same script described in the 
next section (see Sec.\ref{ata}). Each
 curve correspond to a different network condition based on packet drop 
probability. As expected, with more packet dropped, the swarm 
takes more time to reach consensus. The barrier functions described above are used for instance to ensure that
 all robots went through the taking off routine, thus is 
the time taken by the first steps of Fig.~\ref{fig:joinsim}. Up to 90\% of packet drop, the consensus was always reached in simulation.
 
\section{Field Experiments}

Several laboratories are conducting outdoor swarm
experiments~\cite{Davis2016,Hauert2011} and can benefit from our implementation to simplify the development of distributed behaviors. As stated above, this work aims at being
platform-agnostic. It is required to adapt to the specific
needs of the users (developers) and to the growth of the
commercial UAVs market. The infrastructure presented in
Fig. \ref{fig:struc}, was implemented on NVidia TK1 and TX1,
both running Ubuntu to control DJI Matrice 100 quadcopters (M100),
equipped with a Zenmuse X3 camera and a collision avoidance
module (Guidance). It was also ported to a Raspberry Pi 3,
running Raspbian, mounted to the bottom of a 3DR Solo quadcopter,
equipped with a GoPro. The latest addition to this heterogeneous
fleet was the Intel Aero, running the Yocto distribution and
equipped with a Realsense. As proven by the ArduPilot community, the MAVlink
protocol is also perfectly fitted to command and monitor 
rovers\footnote{\url{http://ardupilot.org/rover/}}. ROSBuzz was thus ported to a Clearpath Husky to control its navigation
within a swarm of combined heterogeneous UAVs and UGVs. All robots are equipped with a Xbee 900 Pro communication module for inter-robot exchanges.

Experiments were conducted in an outdoor field with backup pilots
for each UAVs and UGVs. The focus was to validate the infrastructure and specifically the consensus strategy implemented in Buzz. We tested with a behavior attributing tasks in a distributed fashion to the robots~\cite{liAU2017}.

\subsection{Acyclic task allocation}
\label{ata}
A common scenario for a robot group is to execute a given queue of tasks 
evolving throughout the mission. Before optimizing the allocation of the tasks, 
the swarm must have a mechanism to ensure it will reach consensus on a given 
set of allocations. For simplicity, lets represent the tasks with target 
positions in order to form a given graph. It is assumed that all
robots involved in the formation are aware of the graph topology. This is 
achieved by sharing the graph structure table before the robots' deployment or through run-time broadcast. This table contains spatial coordinates of each the nodes, i.e. label representing a task to be assigned to a robot. However, robots 
are not pre-assigned to a specified label (task) in
the graph. The behavior law allows them to find proper labels through simple
local interactions with other robots, including robots already part
of the formation and robots not yet in the formation. This process can drive 
free robots to participate in the
formation gradually or, from the perspective of the formation, it will
attract free robots to join from the edges of the current formation, allowing 
it to grow dynamically.

\begin{figure}[!h]
\centering
\resizebox{6.5cm}{3cm}{
\begin{tikzpicture}
\footnotesize
\begin{scope}[every node/.style={circle,thick,draw, inner sep=0pt}]
    \node [minimum size=12mm] (A) at (-2, 0) {Turned Off};
    \node [minimum size=12mm] (B) at (0, 0) {Take Off};
    \node [minimum size=12mm] (C) at (2, 0) {Free};
    \node [minimum size=12mm] (D) at (4, -1) {Asking};
    \node [minimum size=12mm] (E) at (2, -2) {Joining};
    \node [minimum size=12mm] (F) at (0, -2) {Joined};
    \node [minimum size=12mm, fill=gray!30] (G) at (-2,-2) {Lock};
\end{scope}
   \barrier{0.7,0.6}
   \barrier{-1.3,-1.4}
      \path [-latex',line width=.3mm] (A) edge[bend left] (B);
       \path [-latex',line width=.3mm] (B) edge[bend left]  (C);
       \path [-latex',line width=.3mm] (C) edge[bend left]  (D);
       \path [-latex',line width=.3mm] (D) edge[bend left]  (C);
    \path [-latex',line width=.3mm] (D) edge[bend left]   (E);
    \path [-latex',line width=.3mm] (E) edge[bend left]  (F);
    \path [-latex',line width=.3mm] (F) edge[bend right]  (G);
    
\end{tikzpicture}
}
\caption{The behavior law represented as a finite state machine.
Every robot joining the formation will experience states \emph{TurnedOff}, 
\emph{TakeOff},
\emph{Free}, \emph{Asking}, \emph{Joining} and \emph{Joined}. Before switching 
to state \emph{Free} and \emph{Lock} the robots wait in a transition barrier 
state.}
\label{fig:behavior}
\end{figure}
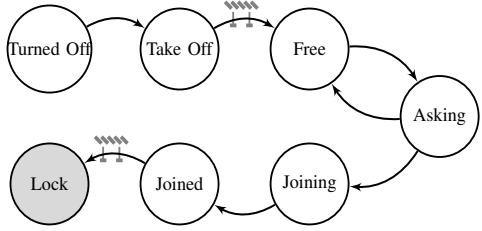

The formation process starts when a robot gets the label 0 in the graph. The progressive attribution of tasks will start from this robot, called the root, it is 
thus considered joined in the formation as soon as it goes out of the barrier after \emph{TakeOff}.
The behavior law is represented as a finite 
state machine, shown in Fig. \ref{fig:behavior}. It consists of seven states: \emph{Turned Off}, \emph{Take Off}, \emph{Free}, \emph{Asking}, \emph{Joining}, \emph{Joined} and \emph{Lock}. After a user sent asked to start the mission, the stakeholder, i.e. the drone the user is connected to, share the information for take off. The assignment of tasks will start only after the first barrier, waiting for all members to be at a safe height. In state \emph{Free}, the robot will circle around the edge of the formation, namely the structure composed of 
\emph{Joining} and \emph{Joined} robots, and search for a proper label in the 
graph. When such a label is found, and both predecessors are within sight, the 
\emph{Free} robot will transit to state \emph{Asking}, sending a message to 
request for the label. Once the request is approved by the \emph{Joining} and 
\emph{Joined} robots, the robot transits to state \emph{Joining}. From that point on 
it is part of the formation and is attributed a position in the graph. With the knowledge of its 
\emph{Joined} parent and of its own label position in the graph, the robot will 
compute its target GPS coordinates and navigate to it. Furthermore, since each robot needs only one predecessor (a robot already joined in the tree), it is not necessary to keep the entire structure of the graph, but rather only a predecessor tree.

\subsection{Observations}
\begin{figure}[!h]
\begin{center}
  \includegraphics[width=0.45\textwidth]{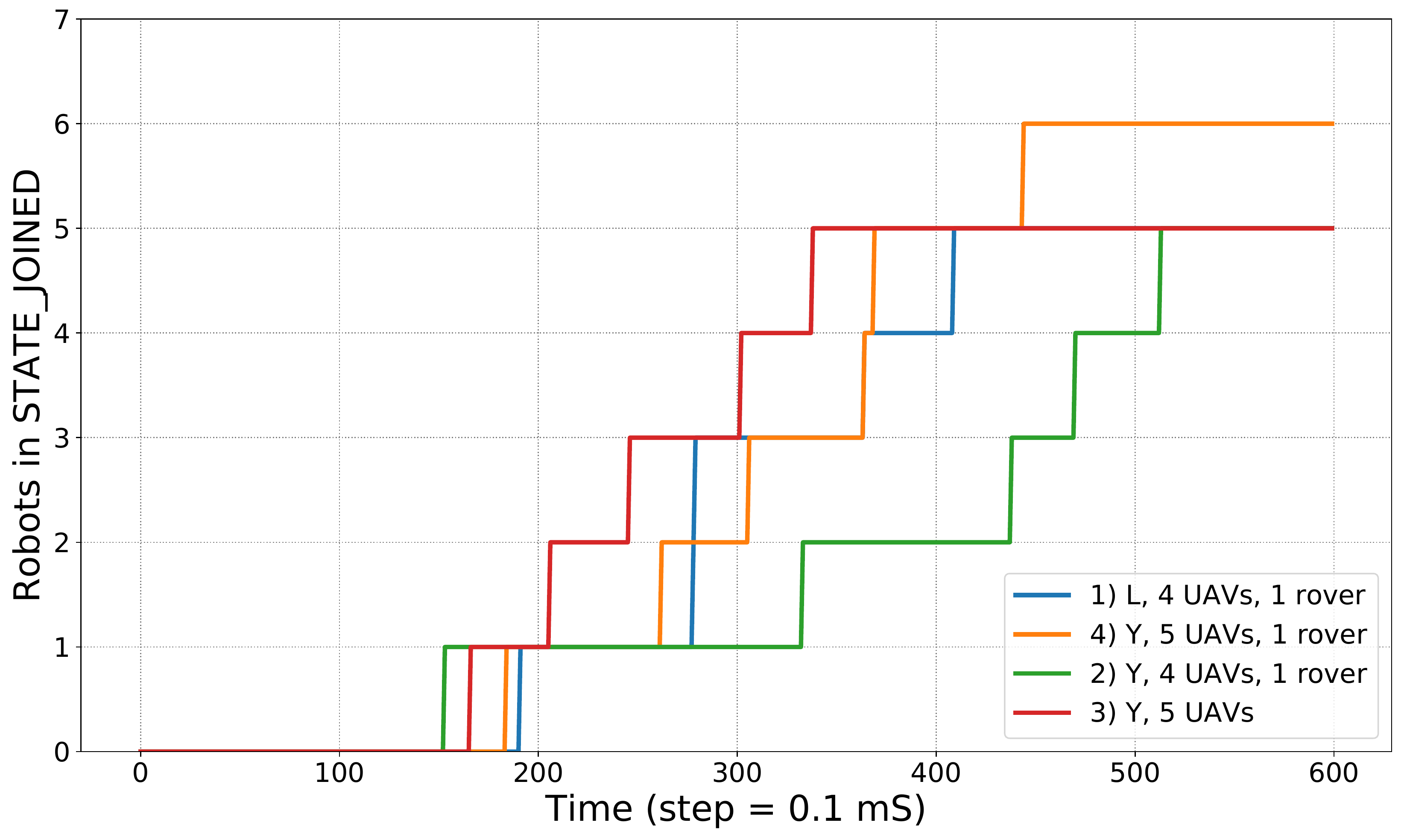}
\caption{Time required for all the robots to join on each of the four experiments conducted. Four experiments had five robot while the orange one had six.}
\label{fig:join}       
\end{center}
\vspace{-1.5em}
\end{figure}

As explained in~\cite{liAU2017} any graph can be generated following the number of nodes (robots) available and a given 2D point cloud (expected geometry). Four experiments were conducted in the field, to test different topology and geometry:
\begin{enumerate}
\item a graph stretching two branches (`L' shape) with 4 M100 and a Husky
\item a graph stretching three branches (`Y' shape) with 5 M100 and a Husky
\item a graph stretching three branches (`Y' shape) with 4 M100 and a Husky
\item a graph stretching three branches (`Y' shape) with 5 M100 only
\end{enumerate}
The time required for each unit to joined the graph, i.e. to get its assigned 
label and move to its target position, is illustrated in Fig.~\ref{fig:join}. 
The first robot to join takes more than 150s because it needs to wait for the whole 
fleet to takeoff and get over the first barrier. As seen in the first experiment, some robots are parents to more than one other and so it is possible to have two robots simultaneously joining the formation. In the last experiment, the first three robots joined in less than 250s most likely because the ground-to-air communication in the other scenarios is slowing down the attribution of the tasks. Except for the third experiment, the average time to get a new robot to join is less than half a minute. By comparing this plot with 
Fig.~\ref{fig:joinsim}, we can estimate the real packet drop probability to 
something between 75\% and 90\%. This pessimist value is based on a uniform, 
constant dropping probability, which do not occur in real experiments.

\begin{figure}[!h]
\begin{center}
  \includegraphics[width=0.45\textwidth]{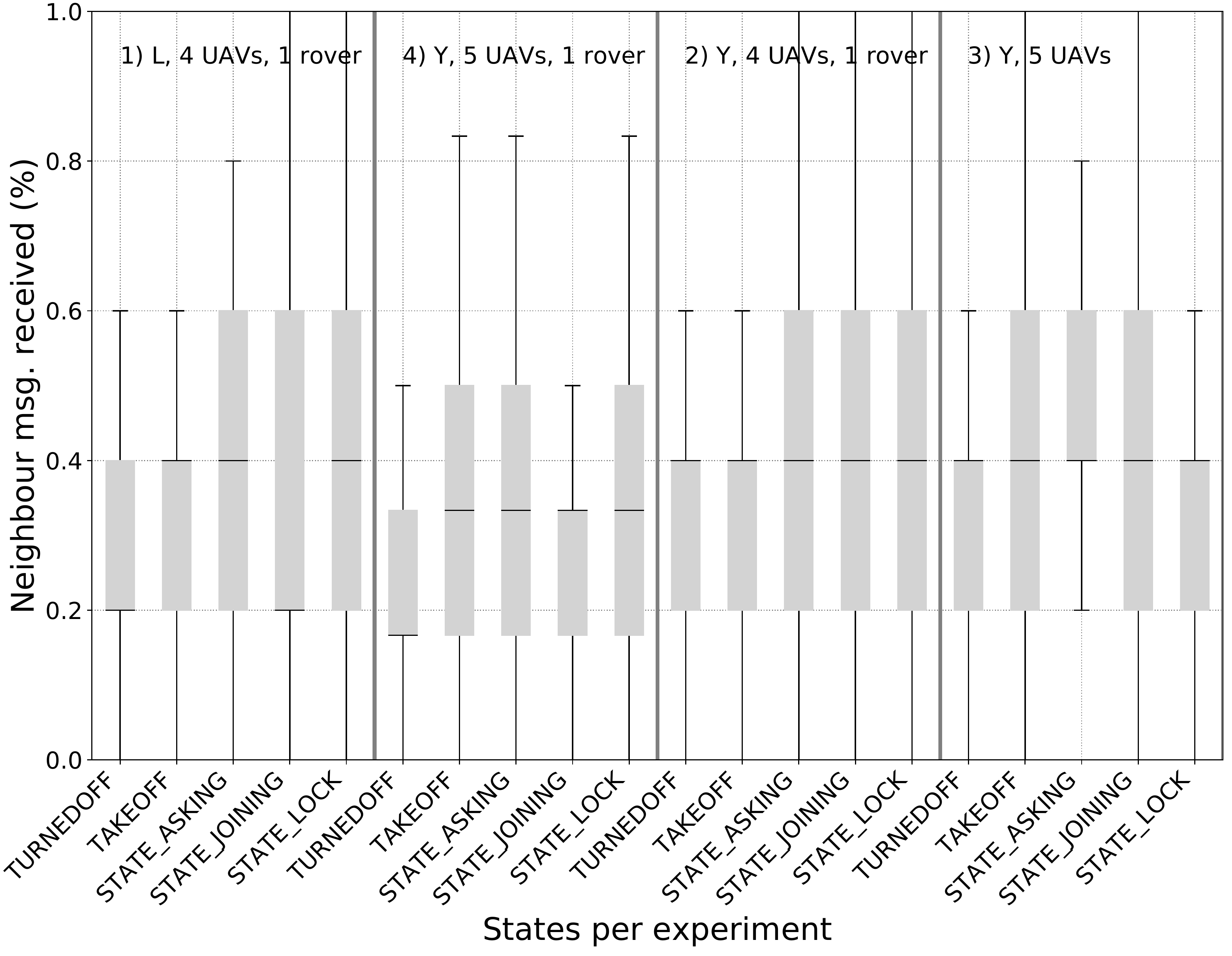}
\caption{Average ratio of neighbors messages received over the total swarm 
member for different states.}
\label{fig:neimsg}       
\end{center}
\vspace{-1.5em}
\end{figure}
The time required to join is influenced by the network performance since each 
robot need to be assigned a label from its parent before moving. With Xbee 
900MHz, the range is large, but the low bandwidth and the packets dropped can 
affect the performance. Fig.~\ref{fig:neimsg} shows the ratio of neighbor 
messages received over the swarm size. Indeed, in a Buzz step, each robot sends 
a message to all its neighbors sharing its position together with a payload 
relevant to the current step operations. We can observe that in average the 
\emph{Turned Off} and \emph{Take Off} states catches less messages than the other states. This can be
explained with the radio wave deflection created by the irregularities of 
the ground.

\begin{figure}[!h]
\begin{center}
  \includegraphics[width=0.4\textwidth]{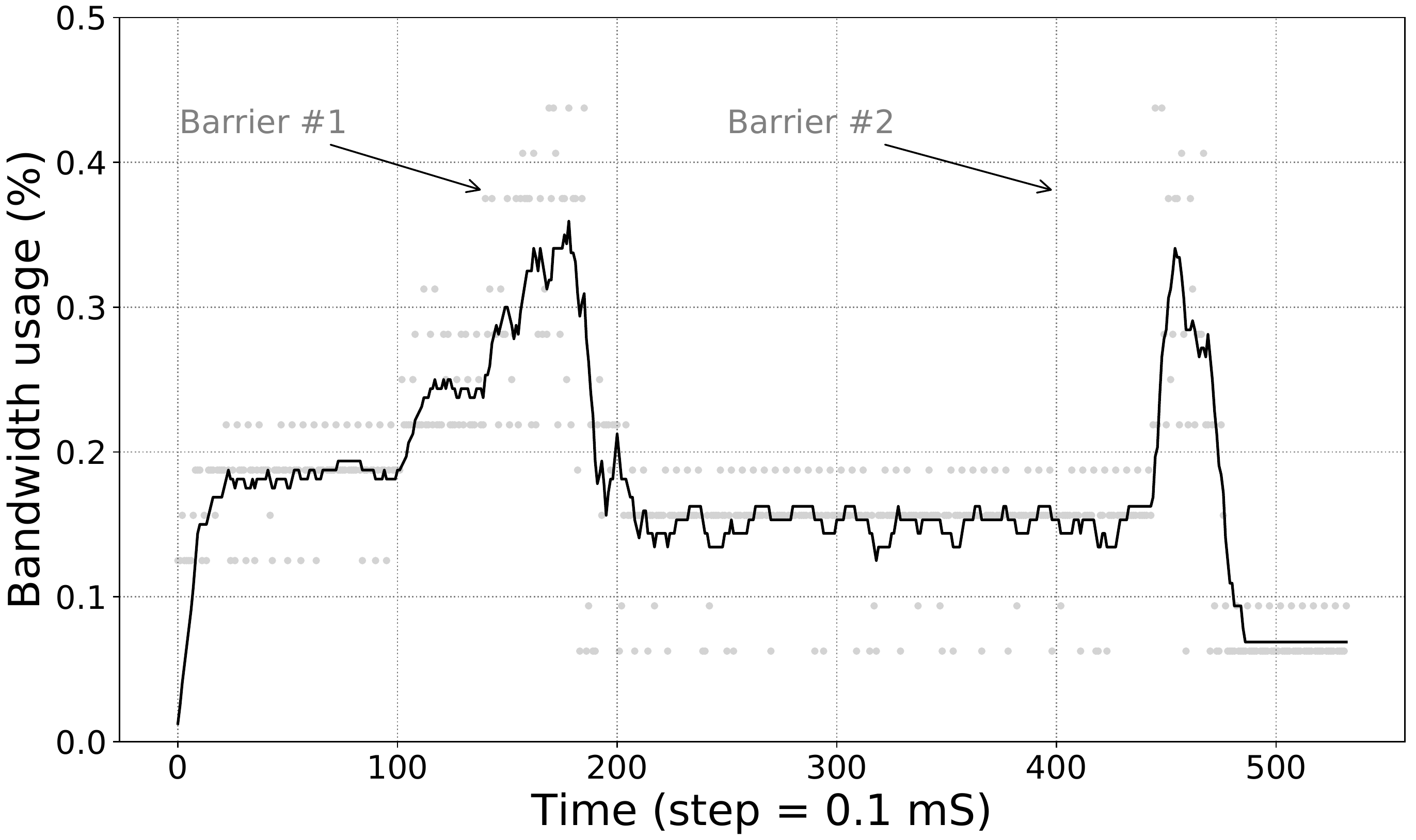}
\caption{Moving average of the bandwidth usage based on a window of 30 samples (which are represented by gray dots).}
\label{fig:band}       
\end{center}
\vspace{-1.5em}
\end{figure}

Finally, Fig.~\ref{fig:band} shows the worst example of bandwidth usage for all 
robots on all experiments. It is clear that the maximum available payload per 
step, i.e. the Xbee frame size (250B, illustrated as a ratio), is never 
exceeded. To better visualize the experiments a video is attached to this paper 
and is available online at \textcolor{red}{mistlab.ca/...}

To better understand the experiments a video is submitted with this paper 
and is available online (\url{youtube.com/mistlab/...}).

\section{Conclusion}

This paper described the advantages of using the software ROSBuzz for the deployment of consensus-based behaviors on multi-robots systems. ROSBuzz is the integration of the swarm-oriented programming language and its virtual machine Buzz into the ROS environment. It grants the developer of distributed behaviors with essential swarm programming premises. As an example, the implementation of a barrier state for the whole group to reach a consensus was detailed. Simulations shown its implementation to be robust to up to 90\% packet drop rate. In order to test the concept and the whole platform-agnostic infrastructure, experiments with a Husky and a fleet of DJI Matrices 100 were conducted in the field. The robots succeeded in each scenario to reach their tasks, i.e. their target positions from a distributed and acyclic assignment mechanism. Through the whole mission, robots used less than half the available bandwidth of the Xbee inter-robot communication modules.

With the experience and results of our field experiments we are optimistic about pushing ROSBuzz to the robotics community. It is already
openly available\footnote{\url{https://github.com/MISTLab/ROSBuzz/}}, just as the scripts described in this paper. More laboratories in Europe and North America
have started using Buzz in their set of software tools and the
community will only continue to grow. The proposed presentation
will cover the swarm basic programming principles, their use in
a Buzz script and how to interface ROSBuzz and XbeeMAV node with
a given robot. As more research will be conducted with this
infrastructure, Buzz and its ROS implementation will be enhanced
and become the solution for swarm intelligence deployment on
robots.

\addtolength{\textheight}{-12cm} 
lengths
last page by a suitable amount.
next page
do not shorten the textheight too much.

%


\bibliographystyle{IEEEtran}
\bibliography{IEEEabrv,swarm,other,references,hri}

\begin{thebibliography}{10}
\providecommand{\url}[1]{#1}
\csname url@rmstyle\endcsname
\providecommand{\newblock}{\relax}
\providecommand{\bibinfo}[2]{#2}
\providecommand\BIBentrySTDinterwordspacing{\spaceskip=0pt\relax}
\providecommand\BIBentryALTinterwordstretchfactor{4}
\providecommand\BIBentryALTinterwordspacing{\spaceskip=\fontdimen2\font plus
\BIBentryALTinterwordstretchfactor\fontdimen3\font minus
  \fontdimen4\font\relax}
\providecommand\BIBforeignlanguage[2]{{%
\expandafter\ifx\csname l@#1\endcsname\relax
\typeout{** WARNING: IEEEtran.bst: No hyphenation pattern has been}%
\typeout{** loaded for the language `#1'. Using the pattern for}%
\typeout{** the default language instead.}%
\else
\language=\csname l@#1\endcsname
\fi
#2}}

\bibitem{Dudek2000}
G.~Dudek and E.~E. Milios, ``{Multi-Robot Collaboration for Robust
  Exploration},'' no. April, pp. 64--69, 2000.

\bibitem{Sahin2004}
E.~$\backslash$cSahin, ``{Swarm robotics: From sources of inspiration to
  domains of application},'' \emph{Swarm robotics}, pp. 10--20, 2004.

\bibitem{kteam}
K-Team, ``{\url https://www.k-team.com/mobile-robotics-products},'' Last
  visited 06/2017.

\bibitem{Goc2016}
M.~L. Goc, L.~H. Kim, A.~Parsaei, J.-d. Fekete, P.~Dragicevic, and S.~Follmer,
  ``{Zooids : Building Blocks for Swarm User Interfaces},'' in \emph{UIST},
  Tokyo, 2016.

\bibitem{Pinciroli2015}
C.~Pinciroli, A.~Lee-Brown, and G.~Beltrame, ``{Buzz: An Extensible Programming
  Language for Self-Organizing Heterogeneous Robot Swarms},''
  \emph{arXiv:1507.05946}, p.~12, 2015.

\bibitem{Brambilla2013}
M.~Brambilla, E.~Ferrante, M.~Birattari, and M.~Dorigo, ``{Swarm robotics: A
  review from the swarm engineering perspective},'' \emph{Swarm Intelligence},
  vol.~7, no.~1, pp. 1--41, 2013.

\bibitem{Hauert2011}
S.~Hauert, S.~Leven, M.~Varga, F.~Ruini, A.~Cangelosi, J.~C. Zufferey, and
  D.~Floreano, ``{Reynolds flocking in reality with fixed-wing robots:
  Communication range vs. maximum turning rate},'' \emph{IEEE International
  Conference on Intelligent Robots and Systems}, pp. 5015--5020, 2011.

\bibitem{reynolds1987flocks}
C.~W. Reynolds, ``Flocks, herds and schools: A distributed behavioral model,''
  \emph{ACM SIGGRAPH computer graphics}, vol.~21, no.~4, pp. 25--34, 1987.

\bibitem{Bayindir2016}
L.~Bayindir, ``{A review of swarm robotics tasks},'' \emph{Neurocomputing},
  vol. 172, pp. 292--321, 2016.

\bibitem{Bamberger2006}
R.~J. Bamberger, D.~P. Watson, D.~H. Scheidt, and K.~L. Moore, ``{Flight
  demonstrations of unmanned aerial vehicle swarming concepts},'' \emph{Johns
  Hopkins APL Technical Digest (Applied Physics Laboratory)}, vol.~27, no.~1,
  pp. 41--55, 2006.

\bibitem{Brunet2008}
L.~Brunet, H.-L. Choi, and J.~P. How, ``{Consensus-based auction approaches for
  decentralized task assignment},'' in \emph{AIAA Guidance, Navigation, and
  Control Conference}, no. August, 2008, pp. 1--24.

\bibitem{WeiRen2005}
{Wei Ren}, R.~Beard, and E.~Atkins, ``{A survey of consensus problems in
  multi-agent coordination},'' in \emph{Proceedings of the American Control
  Conference}, 2005, pp. 1859--1864.

\bibitem{Kuriki2015}
Y.~Kuriki and T.~Namerikawa, ``{Experimental Validation of Cooperative
  Formation Control with Collision Avoidance for a Multi-UAV System},'' in
  \emph{Proceedings of the 6th International Conference on Automation, Robotics
  and Applications}, 2015, pp. 531--536.

\bibitem{Wei2015}
X.~Wei, D.~Fengyang, Z.~Qingjie, Z.~Bing, and S.~Hongchang, ``{A New Fast
  Consensus Algorithm Applied in Rendezvous of},'' in \emph{2015 27th Chinese
  Control and Decision Conference (CCDC)}, 2015, pp. 55--60.

\bibitem{Davis2016}
D.~T. Davis, T.~H. Chung, M.~R. Clement, and M.~A. Day, ``{Consensus-Based Data
  Sharing for Large-Scale Aerial Swarm Coordination in Lossy Communications
  Environments},'' in \emph{IEEE/RSJ International Conference on Intelligent
  Robots and Systems (IROS)}, 2016, pp. 3801--3808.

\bibitem{chung201350}
T.~H. Chung, K.~D. Jones, M.~A. Day, M.~Jones, and M.~Clement, ``50 vs. 50 by
  2015: Swarm vs. swarm uav live-fly competition at the naval postgraduate
  school,'' 2013.

\bibitem{mavlinkweb}
QGroundControl, ``{MAVlink Micro Air Vehicule Communication Protocol},''
  \url{http://qgroundcontrol.org/mavlink/start}, 2017, [Online; accessed
  28-Jan-2017].

\bibitem{Chung2016}
T.~H. Chung, M.~R. Clement, M.~A. Day, K.~D. Jones, D.~Davis, and M.~Jones,
  ``{Live-fly, large-scale field experimentation for large numbers of
  fixed-wing UAVs},'' \emph{Proceedings - IEEE International Conference on
  Robotics and Automation}, vol. 2016-June, pp. 1255--1262, 2016.

\bibitem{Pickem2016}
D.~Pickem, P.~Glotfelter, L.~Wang, M.~Mote, A.~Ames, E.~Feron, and
  M.~Egerstedt, ``{The Robotarium: A remotely accessible swarm robotics
  research testbed},'' in \emph{Proc. of the International Conference on
  Robotics and Automation}, Stockholm, 2016.

\bibitem{camazine2002self}
S.~Camazine, J.-L. Deneubourg, N.~Franks, J.~Sneyd, G.~Theraulaz, and
  E.~Bonabeau, \emph{Self-organization in biological systems}.\hskip 1em plus
  0.5em minus 0.4em\relax Princeton University Press, 2002.

\bibitem{lamport1978time}
L.~Lamport, ``Time, clocks, and the ordering of events in a distributed
  system,'' \emph{Communications of the ACM}, vol.~21, no.~7, pp. 558--565,
  1978.

\bibitem{Pinciroli2016}
C.~Pinciroli, A.~Lee-Brown, and G.~Beltrame, ``{A Tuple Space for Data Sharing
  in Robot Swarms},'' \emph{Proceedings of the 9th EAI International Conference
  on Bio-inspired Information and Communications Technologies (formerly
  BIONETICS)}, pp. 287--294, 2016.

\bibitem{Pinciroli2016b}
C.~Pinciroli and G.~Beltrame, ``{Swarm-Oriented Programming of Distributed
  Robot Networks},'' \emph{Computer}, vol.~49, no.~12, pp. 32--41, 2016.

\bibitem{Støy2001}
K.~St{\o}y, ``{Using situated communication in distributed autonomous mobile
  robots},'' \emph{Proceedings of the 7th Scandinavian Conference on Artificial
  Intelligence}, pp. 44--52, 2001.

\bibitem{Pinciroli2012}
C.~Pinciroli, V.~Trianni, R.~O'Grady, G.~Pini, A.~Brutschy, M.~Brambilla,
  N.~Mathews, E.~Ferrante, G.~{Di Caro}, F.~Ducatelle, M.~Birattari, L.~M.
  Gambardella, and M.~Dorigo, ``{ARGoS: A modular, parallel, multi-engine
  simulator for multi-robot systems},'' \emph{Swarm Intelligence}, vol.~6,
  no.~4, pp. 271--295, 2012.

\bibitem{Meyer2012}
J.~Meyer, A.~Sendobry, S.~Kohlbrecher, U.~Klingauf, and O.~von Stryk,
  ``{Comprehensive Simulation of Quadrotor UAVs using ROS and Gazebo},'' in
  \emph{Simulation, Modeling, and Programming for Autonomous Robots: Third
  International Conference}, no. November, Tsukuba, 2012.

\bibitem{liAU2017}
G.~Li, D.~St-Onge, C.~Pinciroli, A.~Gasparri, E.~Garone, and G.~Beltrame,
  ``Decentralized progressive shape formation with robot swarms,''
  \emph{Journal of Autonomous Robots}, 2017.

\end{thebibliography}

\end{document}